\documentclass[a4paper,USenglish,cleveref, autoref, thm-restate]{lipics-v2021}
\title{Constraint Model for the Satellite Image Mosaic Selection Problem}
\titlerunning{Satellite Image Mosaic Selection Problem} %TODO optional, please use if title is longer than one line

\author{Manuel Combarro Simón}{University of Luxembourg, Luxembourg \and Interdisciplinary Centre for Security, Reliability and Trust (SnT), Luxembourg}{manuel.combarrosimon@uni.lu}{https://orcid.org/0000-0002-2699-1397}{This work is partially funded by the Luxembourg National Research Fund (FNR)---ASTRAL Project, ref. 17043604, and by the joint research programme UL/SnT-ILNAS on Technical Standardization for Trustworthy ICT, Aerospace, and Construction.}
\author{Pierre Talbot}{University of Luxembourg, Luxembourg \and Interdisciplinary Centre for Security, Reliability and Trust (SnT), Luxembourg}{pierre.talbot@uni.lu}{https://orcid.org/0000-0001-9202-4541}{This work is supported by the FNR---COMOC Project, ref. C21/IS/16101289.}
\author{Grégoire Danoy}{University of Luxembourg, Luxembourg \and Interdisciplinary Centre for Security, Reliability and Trust (SnT), Luxembourg}{gregoire.danoy@uni.lu}{https://orcid.org/0000-0001-9419-4210}{}
\author{Jedrzej Musial}{Poznan University of Technology, Poland}{jedrzej.musial@cs.put.poznan.pl}{https://orcid.org/0000-0003-3018-5010}{This work is funded by the FNR---PolLux program under the SERENITY Project, ref.
C22/IS/17395419}
\author{Mohammed Alswaitti}{University of Luxembourg, Luxembourg \and Interdisciplinary Centre for Security, Reliability and Trust (SnT), Luxembourg}{mohammed.alswaitti@uni.lu}{https://orcid.org/0000-0003-0580-6954}{}
\author{Pascal Bouvry}{University of Luxembourg, Luxembourg \and Interdisciplinary Centre for Security, Reliability and Trust (SnT), Luxembourg}{pascal.bouvry@uni.lu}{https://orcid.org/0000-0001-9338-2834}{}

\authorrunning{M. Combarro Simón, P. Talbot, G. Danoy, J. Musial, M. Alswaitti, and P. Bouvry} %TODO mandatory. First: Use abbreviated first/middle names. Second (only in severe cases): Use first author plus 'et al.'


\Copyright{Manuel Combarro Simón, Pierre Talbot, Grégoire Danoy, Jedrzej Musial, Mohammed Alswaitti, and Pascal Bouvry} %TODO mandatory, please use full first names. LIPIcs license is "CC-BY";  http://creativecommons.org/licenses/by/3.0/

 %TODO mandatory; please add comma-separated list of keywords

\category{Short Paper} %optional, e.g. invited paper

\supplementdetails[subcategory={Source Code}, cite={}, swhid={swh:1:dir:c776d54cd26c5685fd8a0a8ddfc2131771884b32;origin=https://github.com/mancs20/mosaic_image_combination;visit=swh:1:snp:dcea6c87bfaa8732f47e6dbcae97b1f0366970a2;anchor=swh:1:rev:68318fe63b0d4986a2c7e6e739d793da93a4979b}]{Software}{https://github.com/mancs20/mosaic_image_combination/tree/cp2023} %linktext, cite, and subcategory are optional

\nolinenumbers %uncomment to disable line numbering

\EventEditors{Roland H. C. Yap}
\EventNoEds{1}
\EventLongTitle{29th International Conference on Principles and Practice of Constraint Programming (CP 2023)}
\EventShortTitle{CP 2023}
\EventAcronym{CP}
\EventYear{2023}
\EventDate{August 27--31, 2023}
\EventLocation{Toronto, Canada}
\EventLogo{}
\SeriesVolume{280}
\ArticleNo{23}
\usepackage[dvipsnames]{xcolor}
\usepackage{soul}
\usepackage{url}
\usepackage[utf8]{inputenc}
\usepackage{graphicx}
\usepackage{amsmath}
\usepackage{amsthm}
\usepackage{amsfonts}
\usepackage{booktabs}
\usepackage{algpseudocode}
\usepackage{listings}
\usepackage{pgfplots}
\usepackage{float}
\pgfplotsset{width=10cm,compat=1.18}
\definecolor{lightgray}{rgb}{0.97, 0.97, 0.97}

\lstdefinelanguage{minizinc}{
    morekeywords={
        ann, annotation, any, array, assert,
        bool,
        constraint,
        else, elseif, endif, enum, exists,
        float, forall, function,
        if, in, include, int,
        list,
        minimize, maximize,
        of, op, output,
        par, predicate,
        record,
        set, solve, string,
        test, then, tuple, type,
        var,
        where,
        abort, abs, acosh, array_intersect, array_union,
        array1d, array2d, array3d, array4d, array5d, array6d, asin, assert, atan,
        bool2int,
        card, ceil, combinator, concat, cos, cosh,
        dom, dom_array, dom_size, dominance,
        exp,
        fix, floor,
        index_set, index_set_1of2, index_set_2of2, index_set_1of3, index_set_2of3, index_set_3of3,
        int2float, is_fixed,
        join,
        lb, lb_array, length, let, ln, log, log2, log10,
        min, max,
        pow, product,
        round,
        set2array, show, show_int, show_float, sin, sinh, sqrt, sum,
        tan, tanh, trace,
        ub, and ub_array,
        bool_search, int_search, seq_search, priority_search,
        minisearch, search, while, repeat, next, commit, print, post, sol, scope, time_limit, break, fail
    },
    sensitive=true, % are the keywords case sensitive
    morecomment=[l][\em\color{blue}]{\%},
    morestring=[b]",
}

\begin{document}

\maketitle

\begin{abstract}
Satellite imagery solutions are widely used to study and monitor different regions of the Earth.
However, a single satellite image can cover only a limited area.
In cases where a larger area of interest is studied, several images must be stitched together to create a single larger image, called a mosaic, that can cover the area.
Today, with the increasing number of satellite images available for commercial use, selecting the images to build the mosaic is challenging, especially when the user wants to optimize one or more parameters, such as the total cost and the cloud coverage percentage in the mosaic. More precisely, for this problem the input is an area of interest, several satellite images intersecting the area, a list of requirements relative to the image and the mosaic, such as cloud coverage percentage, image resolution, and a list of objectives to optimize.
We contribute to the constraint and mixed integer lineal programming formulation of this new problem, which we call the \textit{satellite image mosaic selection problem}, which is a multi-objective extension of the polygon cover problem.
We propose a dataset of realistic and challenging instances, where the images were captured by the satellite constellations SPOT, Pléiades and Pléiades Neo. We evaluate and compare the two proposed models and show their efficiency for large instances, up to 200 images.

\end{abstract}

\section{Introduction}

The space industry is continuously growing and is no longer an exclusive market for military and government applications. According to the most recent report of the European Union Agency for the Space Programme (EUSPA)~\cite{EUSPA2022EUSPAReport}, the global market for navigation systems and earth observation (EO) had revenues of around \texteuro200 billion in 2022 and is expected to reach \texteuro500 billion by 2031.
As access to space has become cheaper, an increasing number of private companies have entered the space business.
Due to advances in satellite design and high-resolution remote sensors, the number of satellite launches dedicated to EO in 2021 was greater than the sum of launches between 2012 and 2016~\cite{UnionofConcernedScientistsUCSDatabase}.
In 2020 more than 100 terabytes of satellite imagery was generated per day~\cite{Mohney2020TerabytesCenters}.

There are several EO-based applications that analyze a vast area of interest (AOI) that can only be covered by combining several adjacent images into a larger one, called a mosaic. Mosaics are crucial for applications such as crop classification~\cite{Hall2018ClassificationImagery,Felegari2021IntegrationMapping}, environmental monitoring~\cite{Nikolakopoulos2018SynergisticGreece,Flood2019UsingAustralia}, and urban development analysis~\cite{Wang2021UrbanData,Piaggesi2019PredictingImagery}.
The mosaicking of satellite images is a complicated process that presents challenges, such as color balancing~\cite{Yu2016ColourLibrary} and image stitching~\cite{Megha2021AutomaticFeature}.

\begin{figure}
    \centering
    \includegraphics[width=\linewidth]{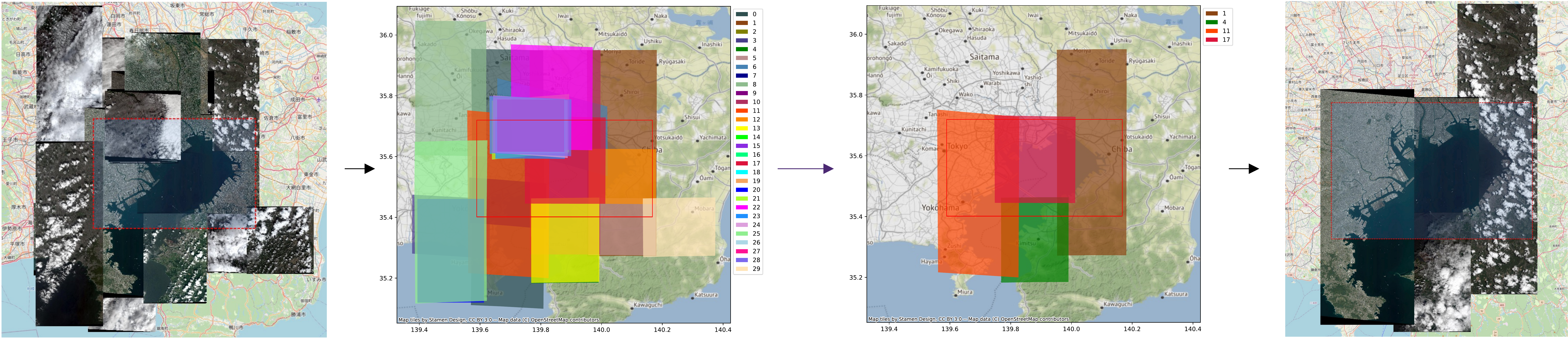}
    \caption{Optimization of the subset of images to make a mosaic of the Tokyo Bay region.}
    \label{fig:satellite_image_selection_problem_described}
\end{figure}

In this work, we focus on the combinatorial problem of selecting the images to create the mosaic by optimizing one or several criteria. This problem is an extended version of the NP-hard problem of finding the minimum axis-parallel rectangle cover of a rectilinear polygon without holes ~\cite{Culberson1988CoveringHard}, where the axis-parallel rectangles can be seen as the satellite images, and the rectilinear polygon as the AOI. In our problem, a cover is the subset of images that can be used to generate a mosaic. In Figure~\ref{fig:satellite_image_selection_problem_described} a particular example of this problem is shown, where the objective is to build a mosaic using the smaller number of images. There are 30 images to choose from, and the optimization algorithm finds an optimal subset of four images. 

In this paper, we present a multi-objective approach for this problem that seeks to optimize four popular parameters of satellite images for mosaic generation: cloud cover, incidence angle~\cite{IncidenceAngle}, resolution and cost of the images. In general, there might not be an objective that is more important than the other, which is why we propose a multi-objective approach, instead of a linear aggregation or lexicographic ordering of the objectives.

As the number of available satellite images has increased significantly, it is becoming more challenging to select the optimal combination of images to build a mosaic. The number of images covering one place can reach hundreds. This is even more difficult if the user is interested in optimizing several parameters. Without a computational approach for this, users have to select by hand the images they want for the cover. Providing a Pareto front from which users can choose a cover is crucial to save money and time, considering that high-resolution satellite images are expensive.

In this paper, we propose a constraint programming (CP) model and a mixed integer linear programming (MILP) model to solve the problem presented. However, directly modeling this geometric problem with constraints is challenging, as we would need to encode geometric operations such as union and intersection of polygons.
Instead, we preprocess each instance by computing a discretization where the intersections of all images are first computed (Section~\ref{sect:discretization_chapter}).
We obtain a set of non-overlapping polygons where each polygon is simply an integer and the geometric characteristics can be ignored.
This problem is a multi-objective extension of the well-known \textit{set covering problem}~(Section~\ref{cp-model}).

A unique aspect of this problem is to minimize the cloud coverage of the mosaic, which, in contrast to other objectives, does not have the same value throughout the image.
While any part of the image has the same resolution, not all parts of the images have the same amount of clouds, except for images with 0 or 100\% of cloud coverage.
Because of this particularity of the problem, it is possible to reduce the cloud coverage percentage in the final mosaic by choosing a specific combination of images in such a way that cloudy regions of an image are overlapped by non-cloudy regions of other images.To the best of our knowledge, there is no work taking this into account to reduce the cloud coverage in the final mosaic. We call this problem the \textit{satellite image mosaic selection problem} (SIMS)~(Section~\ref{problem-spec}).

The main contribution of this paper is to introduce the SIMS problem and present a CP model, as well as MILP model that can successfully find solutions to real instances of up to 200 images (Section~\ref{evaluation-sec}).
The constraint and mixed integer programming approaches are part of a larger framework where the images are automatically retrieved from different marketplaces and the solutions found by the solver can be visualized.

Although SIMS can be expressed as a linear problem (Section~\ref{lp-model}), we choose to rely on constraint programming for two reasons: to ease the formulation of the model---in particular, it is convenient to use set variables---and because this problem aims to be extended for new requirements, which can be non-linear. 
The flexibility of the model is of utmost importance in this work, which is why constraint programming is our main choice.

\section{Satellite Image Mosaic Selection Problem}
\label{problem-spec}

The input for the SIMS problem is an area of interest (AOI) on Earth and a set of satellite images that intersect it.
Each image has a cost and a list of parameters including the resolution, incidence angle and cloud coverage.
The AOI is represented as a simple closed polygon without holes, and the images are represented as quadrilaterals.
For both the AOI and satellite images, the corner coordinates are provided. With that information, the AOI and the images can be represented in the plane.

\begin{figure}
    \centering
    \begin{subfigure}{0.45\textwidth}
        \includegraphics[width=0.7\textwidth]{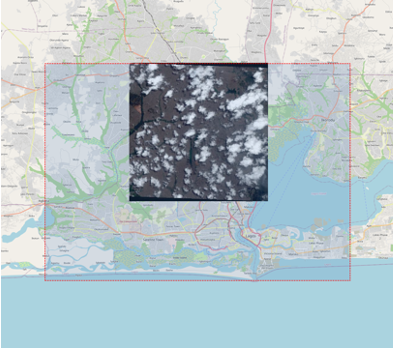}    
    \end{subfigure}
    \begin{subfigure}{0.45\textwidth}
        \includegraphics[width=0.7\textwidth]{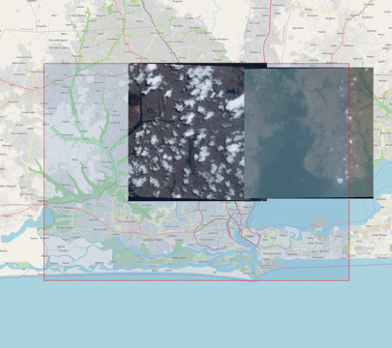}    
    \end{subfigure}
    \caption{A cloudy region of an image can be covered by a non-cloudy region of another image, impacting the cloud coverage percentage of the final mosaic.}
    \label{fig:cloud_effect}
\end{figure}

As clouds are not usually even distributed in the images, having images with a certain cloud coverage percentage does not guarantee that the final mosaic has less than that cloud coverage. Depending on the cloud distribution in the images, the final mosaic can have a lower or higher percentage of cloud coverage as depicted in Figure~\ref{fig:cloud_effect}. This is not the case for the other objectives because they have a unique value along the image.
For example, if all images have a determined resolution, the final mosaic will have the same resolution.

\subsection{Preprocessing of the Problem}\label{sect:discretization_chapter}

\begin{figure}
    \centering
    \includegraphics[width=\linewidth]{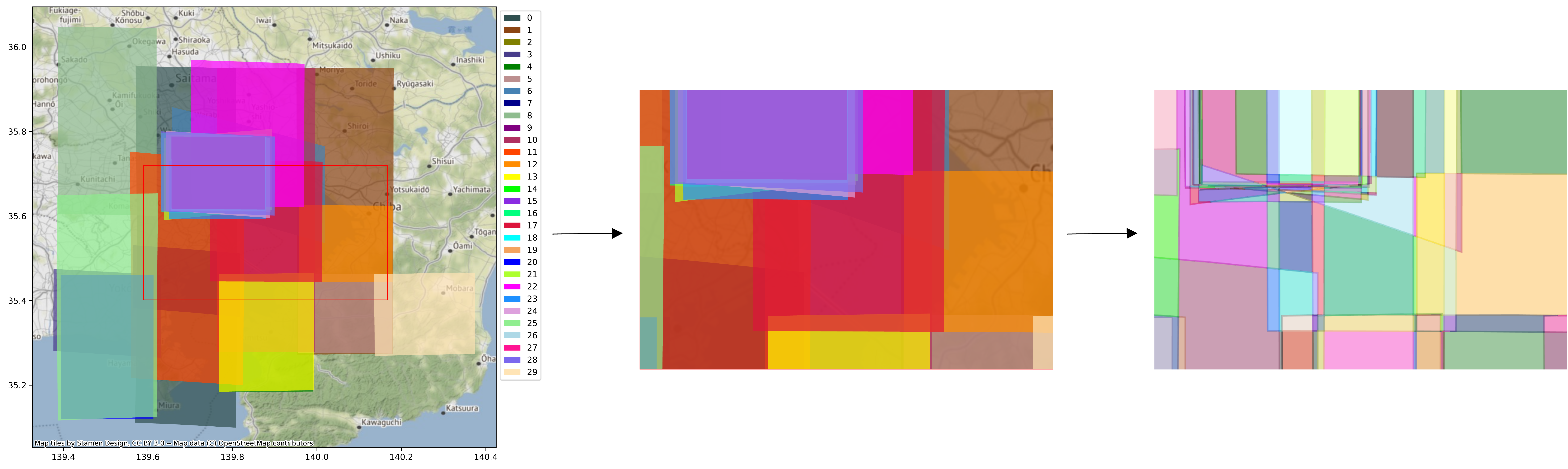}
    \caption{298 polygons are obtained after preprocessing of 30 images. First the area of the images outside the AOI is removed and then the polygons resulting from the images intersections are found.}
    \label{fig:preprocessing_discretize}
\end{figure}

To make the discretization, we first remove the parts of the images that are outside the AOI, and then we find all the polygons resulting from the intersection of the images, we find the polygons using the GEOS library~\cite{GEOSGEOS}.
The universe is partitioned into a set of polygon elements, and each of them is assigned to its corresponding images.
In Figure~\ref{fig:preprocessing_discretize}, we show an example of this process, where we generate 254 polygons from 30 images.
In Table~\ref{interesctions-table}, we show the number of intersections for different cardinality of the images set to cover the Tokyo Bay region.

\begin{table}
    \centering
    \caption{Number of intersections for the instances covering the Tokyo Bay region.}
    \begin{tabular}{c | c c}
    Images & Intersections \\
    \hline
     30 & 298 \\
     50 & 806 \\
     100 & 3278 \\
     150 & 8079 \\
     200 & 14855 \\
    \end{tabular}
    \label{interesctions-table}
\end{table}

The following step is to detect the clouds in the images and add them to the universe and to the correspondig sets. The objective of doing this is to know whether a region of the final mosaic, that is represented by one element of the universe, is free of clouds or not. We consider that a region of the AOI is free of clouds if there is at least one image in the cover in which that region does not have clouds.

A cloudy region is an element in the universe that is present in all the sets that cover that region. For the set that represents the image in which the cloud occurred, we make a distinction and we say that the element is cloudy for that set. In this way, a set is composed of cloudy elements and non-cloudy elements.

In real applications, clouds can be detected using cloud detectors~\cite{Sun2020SatelliteDC,Guo2020CloudDF,Jeppesen2019ACD}.
As this is a problem orthogonal to our work, we do not detect the clouds, but instead randomly allocate them in the parts of the image. For each image, we have metadata indicating the cloud coverage percentage of the image. With that information and knowing the elements that belong to the image, we randomly set one of the elements as cloudy. We repeat this operation until the cloud coverage percentage of the image is achieved. 

\begin{figure}
    \centering
    \begin{subfigure}{0.45\textwidth}
        \includegraphics[width=0.7\linewidth]{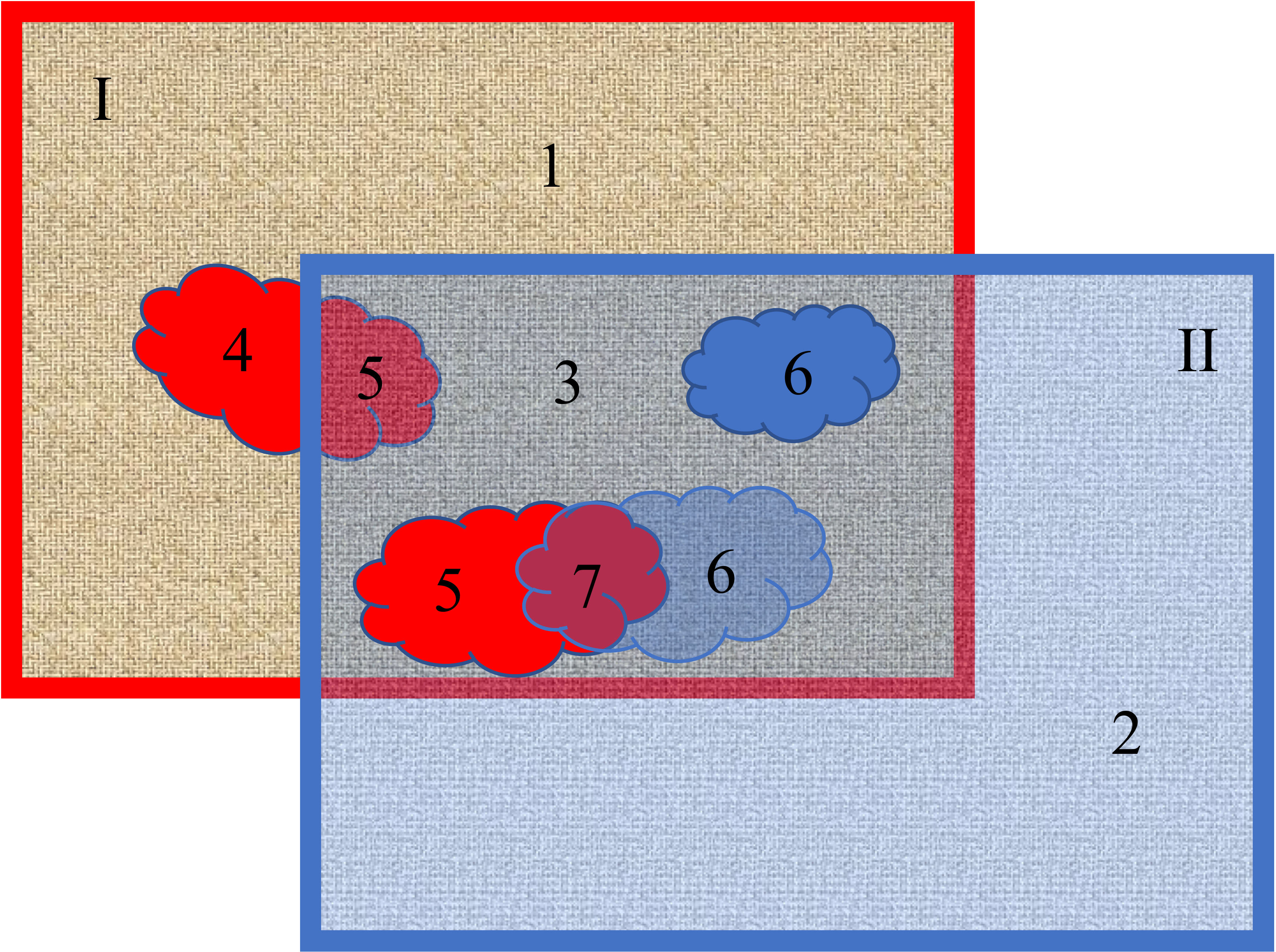}
        \caption{Before the cloud integration the universe had 3 elements and 7 after the integration.}
        \label{fig:cloud_integration_to_universe}
    \end{subfigure}
    \begin{subfigure}{0.45\textwidth}
        \includegraphics[width=0.7\linewidth]{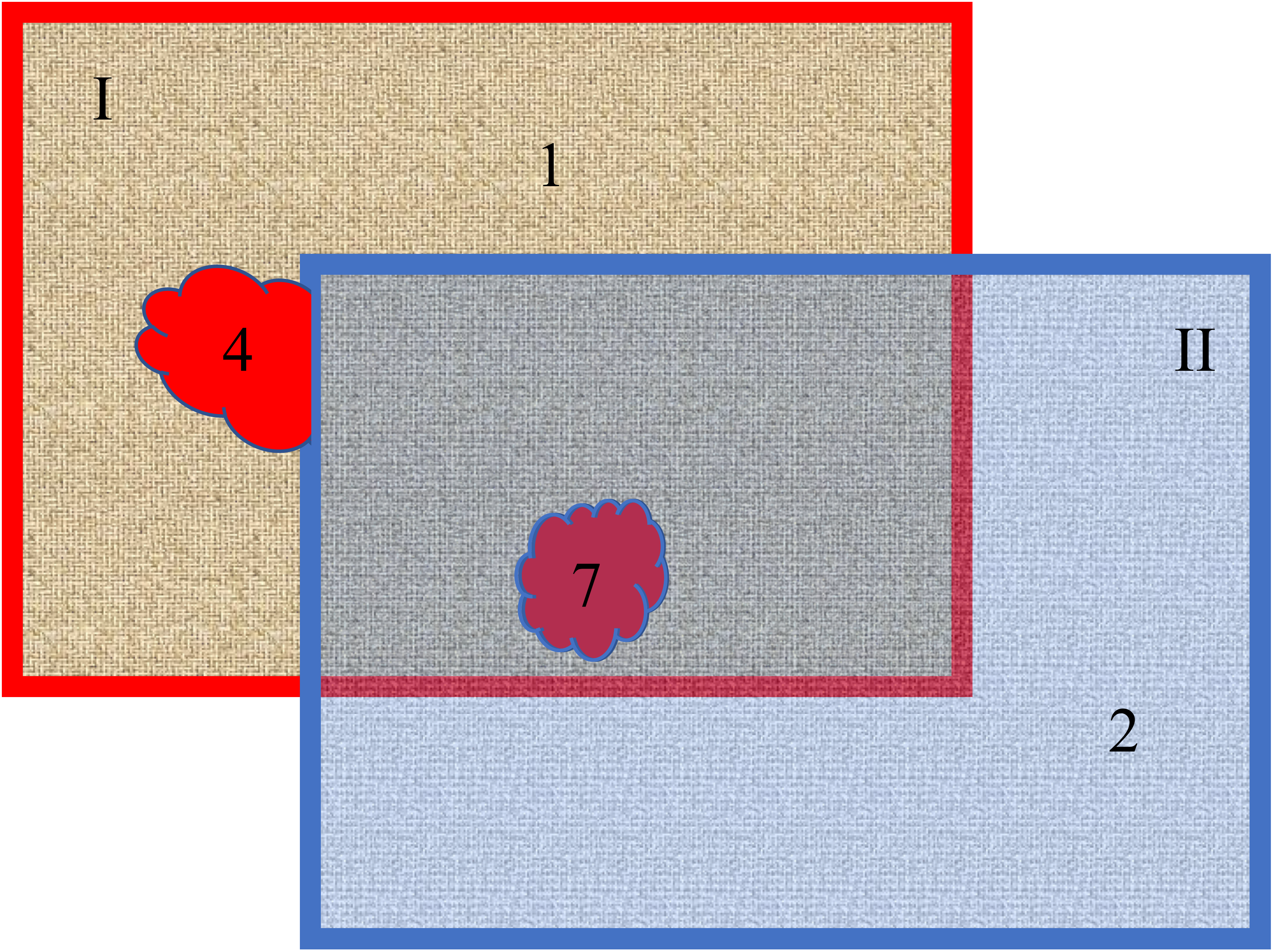}
        \caption{Regions represented by element 5 and 6 are free of clouds in the final mosaic}
        \label{fig:cloud_reduction_after_cloud_integration_to_universe}
    \end{subfigure}
    \caption{Possible cases for the integration of the detected clouds to the universe and the corresponding sets. Set I and II are represented by the rectangles with red and blue borders respectively.}
    \label{fig:cloud_integration_and_reduction}
\end{figure}

In Figure~\ref{fig:cloud_integration_to_universe}, all possible scenarios of how clouds are converted to elements of the universe are shown for the general case. To facilitate the understanding of this process, only two overlapping images are shown, but the procedure is the same when more than two images overlap. Images I and II are represented as rectangles with red and blue borders, respectively. The clouds in image I are colored red, and the ones in image II are colored blue. Initially, both sets have in common its intersection, element 3 ($I = \{1,3\}$ and $II = \{2,3\}$). We can see that both clouds of I are partially covered by II; in one case is because one part of the cloud is in the intersection, and in the other case is because one part of the cloud is overlapped by a cloud from II, so it can not be completely covered by a non-cloudy region of II. From the previous, we can see that three elements are created, 4, 5 and 7. Element 4 are the clouds that are only present in I, element 5 is the cloudy region of I that is not cloudy in II, i.e. covered by II, and element 7 is a cloudy region of I that is also cloudy in II. Element 6 is similar to element 5, is a cloudy area in II that is not cloudy in I. When all the clouds are detected and incorporate to the universe, the original three elements 1, 2 and 3 are modified as follows: element 1 is the non-cloudy region of I that is not overlapped by any other image, element 2 is equivalent to element 1 but for image II, and element 3 is the non-cloudy area of the intersection between I and II. Finally, the universe has seven elements, and the sets are $I=\{1,3,4,5,6,7\}$, $II=\{2,3,5,6,7\}$.
In Figure~\ref{fig:cloud_reduction_after_cloud_integration_to_universe}, we show a resulting mosaic after covering the clouds.
Importantly, taking into account the clouds in this way increases the cardinality of the universe, but the problem itself does not change, which is why we took a simpler approach to randomly assign clouds to each element.

\section{Constraint Model}
\label{cp-model}

Let $U = \{k_1, \ldots, k_n\} \subset \mathbb{N}$ be a set of $n$ parts of the area of interest, called the universe.
The set $U$ is a polygon partition of the area of interest, i.e. two parts do not overlap and their union is exactly the area of interest.
Each satellite image is represented by a collection $P_i \subset U$ of parts.
We write $I = \{P_1, \ldots, P_m\}$ the set of all $m$ satellite images.
The goal is to find a subset $T \subset \{1,\ldots,m\}$ of images that covers the area of interest.
The parameters of the model are $U$ and $I$ while $T$ is the main decision variable.
The set covering constraint is captured by the following:
\begin{equation}
\bigcup_{i \in T} P_i = U
\label{set-cover}
\end{equation}

A trivial solution to this constraint is to take all the images, but we usually consider an optimization version where the cardinality of $T$ is minimized.
In our case, each image $i \in \{1,\ldots,m\}$ has a cost $W_i \in \mathbb{N}$ that we seek to minimize:

\begin{equation}
\min \sum_{i \in T} W_i
\label{cost-constrained}
\end{equation}

\noindent
Along with Equation~\ref{set-cover}, this problem is called \textit{weighted set cover}.
Depending on the user requirements, we can consider other objectives such as resolution and incidence angle.
For each part $k \in U$, we have its area $A_k \in \mathbb{N}$.
And for each image $i \in \{1,\ldots,m\}$, we have its resolution $R_i \in \mathbb{N}$ and its incidence angle $F_i \in \mathbb{N}$.
We seek to minimize (the resolution is given in how many $cm^2$ represents a pixel, the less the better) the best resolution obtained for each part:
\begin{equation}
\min \sum_{k \in U} \min \{ R_i \;|\; i \in T, k \in P_i \}
\label{resolution-part-cp}
\end{equation}
For the incidence angle, we seek to minimize the maximal angle, although other choices would be possible such as minimizing the average.
\begin{equation}
\min~\{ \max~\{ F_i \;|\; i \in T\}\}
\label{incidence-angle-cp}
\end{equation}

A more challenging aspect of this problem is to minimize the area covered by clouds.
To achieve that, we consider that each part is either cloudy or not.
We leave the cloud detection and the splitting of the image into cloudy and non-cloudy parts to a preprocessing step.
Let $C_i \subset P_i$ the cloudy parts of the image $i$.
For each part $k \in U$, we define $D_k := \{ i \in \{1,\ldots,m\} \;|\; k \in P_i \setminus C_i\}$ the set of all images containing a non-cloudy view of the part $k$.
For each part $k \in U$, we can now define the Boolean variable $V_k$ to be true when the part $k$ is cloudy in the cover:
\begin{equation}
V_k \Leftrightarrow \bigwedge_{i \in D_k} i \notin T
\end{equation}
We can now minimize the area covered by clouds:
\begin{equation}
\min \sum_{k \in U} V_k * A_k
\end{equation}

This objective can also be turned into a constraint if the user only wants covers with a certain cloud coverage threshold.

The model introduced is actually linear, as shown in Appendix~\ref{lp-model}, and can be solved by mixed integer programming solvers.
We simply represent the set $T$ by $m$ 0-1 variables $\{x_1, \ldots, x_m\}$ such that $x_i = 1$ if we take the image and $x_i = 0$ otherwise.
We also use this representation for constraint programming solvers, because it is not possible to represent the set covering constraint otherwise---this is due to $T$ having a non-fixed cardinality.

\subsection{Search Strategy Based On Greedy Algorithm}
\label{sect:search_strategy}

A well-known greedy algorithm for the set covering problem consists in taking the images covering the most uncovered parts of the universe first~\cite{Chvatal_1979}.
We model this heuristic as a search strategy within the MiniZinc constraint model.
This has the advantage of always producing a solution that is at least as good as the greedy heuristics---since it is the first solution found.
To achieve that, we reuse an existing search strategy provided by MiniZinc.
A second advantage is that our search strategy can be reused with any constraint solver compatible with MiniZinc.
We select the variable using the \texttt{anti\_first\_fail} strategy---the variable with the largest domain is selected first---and we take the highest value in its domain (\texttt{indomain\_max}).
The trick is to model a set of variables $\{G_1,\ldots,G_m\}$ such that $G_i \in \{0,\ldots,|P_i|\}$ is equal to the number of parts covered by the image $i$.
Actually, in any solution, we have $G_i = |P_i|$ since the whole universe must be covered.
What is interesting is the value of $G_i$ in partial assignments during the search.
The difference between the upper and lower bounds $\mathit{max}(G_i) - \mathit{min}(G_i)$ is the number of parts that are currently uncovered by the partial assignment, and that can be covered by the image $i$.
Since the anti-first-fail strategy selects the largest domain first, it effectively implements the greedy heuristics.
We model $G_i$ as follows:
\begin{equation}
G_i = \sum_{k \in P_i} (\bigvee_{i \in T} k \in P_i)
\end{equation}
\noindent
We note that the new variables $G_i$ are fully defined with the parts, and therefore once the main decision variable $T$ is assigned, the variables $G_i$ must be assigned as well.

\subsection{Multi-Objective Constraint Optimization Algorithm}

The multi-objective constraint programming algorithm used in this work was pionneered by Gavanelli~\cite{gavanelli1_algorithm_2002} and has been frequently used in constraint optimization~\cite{Lukasiewycz-2007,hutchison_multi-objective_2013,guns_solution_2018}.
The main idea is to run a \textit{satisfaction constraint solver} iteratively and add new constraints representing the Pareto front to ensure the next solution is not dominated by any point in the current Pareto front.
To illustrate this algorithm, suppose a biobjective maximization problem where $x$ and $y$ are the two variables to optimize.
We run the constraint solver which returns a first satisfiable solution where $x = 10$ and $y = 5$.
At that point the Pareto front is $\{(10, 5)\}$.
We add to the model the constraint $x > 10 \lor y > 5$ which guarantees that the next solution will not be dominated by the current points in the Pareto front.
The solver might then find the solution $x = 2$ and $y = 6$ which is incomparable to the previous solution and is added to the Pareto front $\{(10, 5), (2, 6)\}$.
The constraint generated from the Pareto front is now $(x > 10 \lor y > 5) \land (x > 2 \lor y > 6)$.
This process continues until the solver finds an unsatisfiable solution, in which case we are guaranteed to have found the optimal Pareto front.

\section{Evaluation}
\label{evaluation-sec}

\subsection{Dataset Description}

For each experiment, there is an AOI and a number of images that cover the AOI. The objective is to find the Pareto front, where each point in the front represents a subset of images that must cover the AOI and optimize four objectives: cost, resolution, incidence angle and cloud coverage.

To carry out this research, we developed a framework capable of retrieving image metadata from different satellite marketplaces, preprocessing it (discretization and cloud integration to the universe), calling a CP or a MILP solver, and visualizing the solutions from the Pareto front.

We selected five AOIs from around the world: Mexico City (Mexico), Rio de Janeiro (Brazil), Paris (France), Lagos (Nigeria), and Tokyo Bay (Japan). For each AOI, we obtained all available images that were captured from 01-01-2021 to 01-01-2023 by the following satellite constellations SPOT~\cite{AirbusSPOT}, Pléiades~\cite{AirbusPleiades} and Pléiades Neo~\cite{AirbusPleiadesNeo}. We opted for those satellite constellations as they have all the metadata used in the experiments; other satellite constellations lacked some parameters such as cloud coverage or incidence angle.

Five instances were generated for all AOIs, except Lagos. Each of these instances differs from each other by the number of images given to cover the AOIs. The number of images for the instances were 30, 50, 100, 150 and 200. For Lagos, the total number of images available for the specified date range was 145, so the number of images for the Lagos instances were 30, 50, 100 and 145.

\subsection{Experimental Setup}

Each instance was solved using the CP model and the MILP model. We run the CP model with two solvers, OR-tools ~\cite{ortools} and Gecode 6.3.0 ~\cite{GeCode}. For each of these solvers, we run the experiments twice; one with the default solver search strategy, and the other one with the greedy search strategy proposed in Section~\ref{sect:search_strategy}. The MILP model was implemented using the Gurobi solver ~\cite{gurobi}. We will refer to these five approaches as \textit{OR-tools default}, \textit{OR-tools greedy}, \textit{Gecode default}, \textit{Gecode greedy} and \textit{Gurobi}

For the MILP model, the algorithm used to obtain the exact Pareto front was SAUGMENCON~\cite{Zhang2014ASA} which is based on the AUGMECON~\cite{MAVROTAS2009455,MAVROTAS20139652} algorithm and on the well-known $\epsilon $-constraint method. In the $\epsilon$-constraint methods, one objective is optimized, and the others are added as constraints to the model. The right-hand side of the objective constraints gradually changes from the less restrictive values of the objectives to the most restricted ones. This process continues until all combinations of values for the constraint objectives have been explored. The SAUGMENCON method introduces two acceleration mechanisms to improve the computational efficiency of the front generation.

The experiments were run on an AMD Epyc ROME 7H12 processor (64 cores, 280W). All the solvers were configured to run in parallel with 8 cores and 16 threads. The running time for each experiment was 1 hour.

\subsection{Experimental Results}

To compare the results, we used the hypervolume of the Pareto front, which is a standard metric for comparing fronts in multiobjective optimization. For each instance, we score the strategies, calculating how worst they are compared to the best. For example, a score of 1 means that the strategy has the same hypervolume value as the best one, and a score of 0.5 means that the hypervolume of the strategy is half of the best hypervolume for that instance.

In Table ~\ref{experimental-result-table} from Appendix ~\ref{experimental-result-detailed} we can see the hypervolume values for each strategy for all instances.

For only one instance, the complete Pareto front was found in the running time, the strategies that found the complete front were \textit{OR-tools default}, \textit{OR-tools greedy} and \textit{Gurobi}. For the rest of the instances, the whole Pareto front was not found, and the hypervolume corresponds to the partial front found during the running time. For 3 of the 4 instances with 200 images, the CP strategies could not find any point of the front; the entire running time was employed by the FlatZinc submodule of Minizinc, to flatten the model with the data file. However, Gurobi for 3 of these instances could find one point of the Pareto front. For the rest of the instances, all the strategies could find at least one point of the Pareto front, and generally the CP approaches obtained superior results compared to MILP.

\begin{figure}
    \centering
    \includegraphics[width=0.49\textwidth]{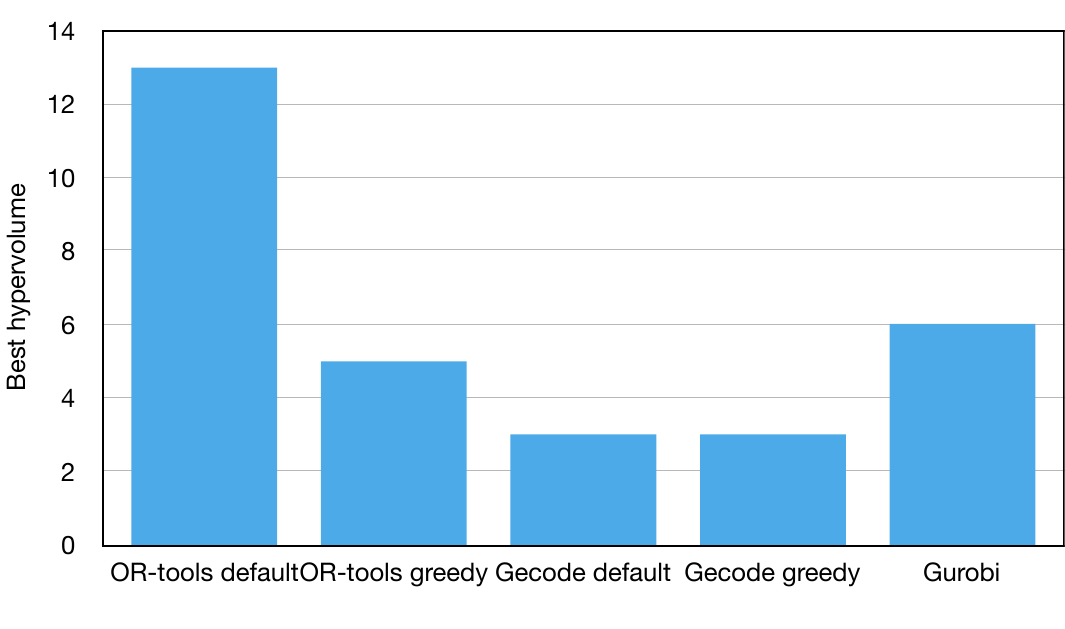}
    \caption{Number of times each approach had the best hypervolume.}
    \label{fig:best_hypervolume}
\end{figure}

\begin{figure}
    \centering
    \begin{subfigure}[t]{0.49\textwidth}
        \includegraphics[width=0.9\linewidth]{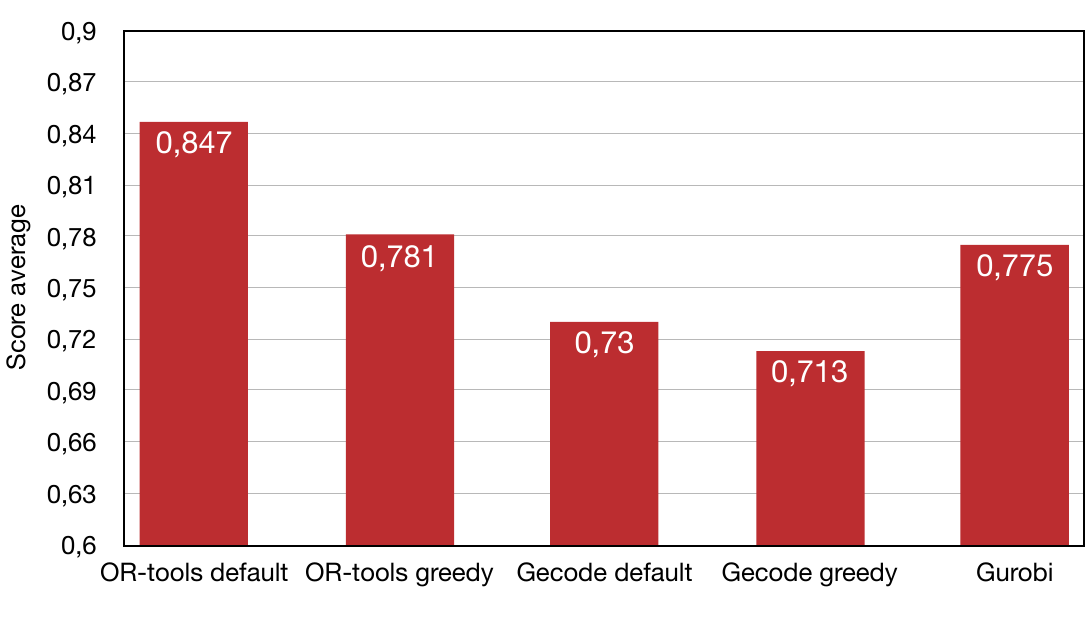}
        \caption{Considering the 24 instances.}
        \label{fig:ave_score_all_experiments}
    \end{subfigure}
    \hfill
    \begin{subfigure}[t]{0.49\textwidth}
        \includegraphics[width=0.9\linewidth]{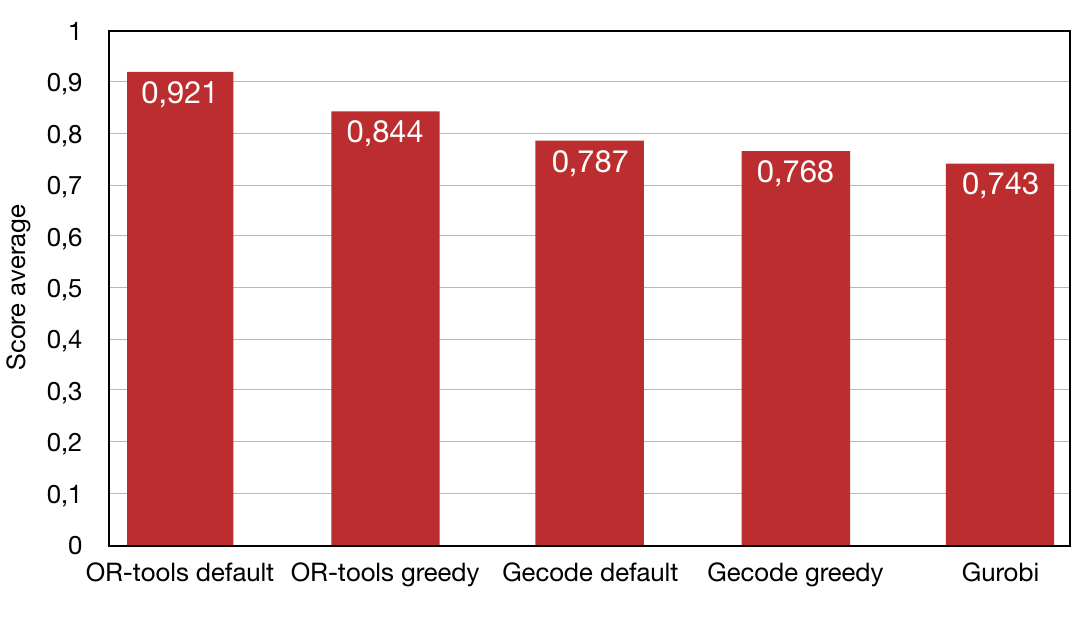}
        \caption{Considering the 21 instances where all strategies found a solution.}
        \label{fig:ave_score_21_experiments}
    \end{subfigure}
    \caption{Average score for each approach.}
    \label{fig:ave_score}
\end{figure}

As we can see in Figures ~\ref{fig:best_hypervolume} and ~\ref{fig:ave_score_all_experiments}, for these experiments, the best approach was \textit{OR-tools default}, being the best for 13 out of 24 instances and with a score average of 0.847. The second and third best strategies were \textit{OR-tools greedy} and \textit{Gurobi}, with a similar performance. The fourth and fifth places were occupied by Gecode default and greedy, being really close.

If we do not consider the 3 instances in which the CP strategies could not find a solution, the score averages change; see Figure ~\ref{fig:ave_score_21_experiments}. \textit{OR-tools default} has an average score very close to 1, and \textit{OR-tools greedy} has a much better average score than \textit{Gurobi}, which for these instances is the worst strategy on average. The difference in the average score between both Gecode strategies remains very close.

It is interesting to note that \textit{Gurobi} showed excellent performance for small instances comprising 30 and 50 images. However, its performance was not as impressive for larger instances. This could be related to the way solutions are discovered and added to the front. For future research, it could be interesting to compare the hypervolume anytime behavior for different approaches used for CP and MILP to get the exact Pareto front.

\section{Related Work}
\label{related-work-sec}
Geometric set covering problems can be divided into two categories based on the requirements of the covering shapes. In one category, the covering shapes do not have a fixed position in the plane, for example, covering a polygonal region with the minimum amount of fixed sized rectangles~\cite{Mansouri2017OnTC} or with a set of known rectangles that can freely move on the plane~\cite{Stoyan2009CoveringByrectangles}. The other category is where the covering shapes have a fixed size and position, for example, covering a polygonal region with discs with a fixed size and position on the plane~\cite{CONTARDO202125}. SIMS, belongs to the second category as the satellite images represent a fixed region on the Earth.

Considering the cloud coverage percentage in the final mosaic makes SIMS problem different from polygon cover and other geometrical cover problems, where the covering shapes only have to cover the polygon or the universe of points in the space.
In this problem, the covering shapes, besides covering the polygon, should also cover certain regions (clouds) that are present in the shapes. Interestingly, this can be seen as solving two weighted set covering problems; in one, the AOI must be covered and in the other, the clouds.

The main approaches to solving geometric set covering problems are local search~\cite{Bansal2012WeightedSampling,De2023GeometricLocal-Search,Ashok2019LocalSS} and linear programming (LP)~\cite{Chan2020FasterCoverb,Chekuri2020FastLA}.
In most of the papers, opposite to SIMS, the universe is a set of points instead of a region.
In~\cite{CONTARDO202125}, they provide an exact algorithm for the case where the universe is a set of regions, and the covering objects are discs. The algorithm is effective when the minimum number of discs to cover the space is low.

In~\cite{Mouthuy2007GlobalProblem}, set covering problem is tackled using constraint programming. There, the authors propose a way to prune the domain of possible solution using a lower and upper bound for the objective value.
The lower bound consists of determining the minimum number of sets that can cover the space.
This is equivalent to answering the following NP-complete problem: does a cover of the universe exist with $K$ sets.
They propose a new strategy to get an approximation of this lower bound and compare it against two other well-known lower-bound values: the value of the LP relaxation problem and a greedy algorithm. The proposed prune strategy is good for problems where the size of the sets is small, for bigger subsets, they recommend alternating between the LP relaxation and the greedy algorithm.

\section{Conclusion}

In this paper, we introduce a novel geometrical NP-hard problem, SIMS, inspired by the selection of satellite images for mosaic generation. CP and MILP models are provided for this problem, together with a search strategy for the CP model, based on the well-known greedy algorithm used for set covering problems. In the experiments performed, the CP solved with OR-tools got the best result, evidencing the power of this solver. Our proposed search strategy could not outperform the default search strategies, but in the case of the Gecode solver, it produced similar results. Generally, the CP model outperformed the MILP model. This could be related to the method used to generate the Pareto front. For future work, it will be interesting to compare different approaches to generate the exact Pareto front for the CP and MILP models, based on the metric anytime behavior for the hypervolume. We also plan to propose heuristics to tackle larger instances and to evaluate their performance against the proposed CP and MILP models.

\bibliography{cp2023}

\appendix

\section{Linear Programming Model}
\label{lp-model}
For the mixed integer linear programming, we use the same nomenclature as for the constraint model. We just add the necessary variables to linearize the model.

To linearize the cover constraint (\ref{set-cover}) it is necessary to associate each image $P_i$ with a decision variable $x_i$ that is equal to 1 if the image $i$ is selected, otherwise it is 0. We rewrite the constraint as follows:

\begin{equation}
    \sum_{i:k \in P_i} x_i \geq 1, \text{for all $k \in U$}
    \label{mip-set-cover}
\end{equation}

The previous constraint guarantees that all the parts are covered by at least one image.

To linearize the cost constraint (\ref{cost-constrained}) it is necessary to associate each image $P_i$ with an auxiliary variable $w_i$ that represents the cost of the image. The linear constraint can be written as:

\begin{equation}
\min \sum_{P_i \in I} x_iw_i
\label{mip-cost}
\end{equation}

The constraints (\ref{mip-set-cover}) and (\ref{mip-cost}) are the classical constraints used for set covering problems.

The resolution objective is a min-min problem, where the objective is to minimize the sum of the min resolution of each part. The min resolution of a part is the minimum resolution of the images that contain them and belong to a cover. We need to add an auxiliary decision variable $r_k$ representing the best resolution of the part $k$ and a big constant $B$, bigger than the maximum image resolution. Also, we need to add an auxiliary binary decision variables $z_{k_j}$ for each image $P_j$ that contains $k$. For each part $k$, we define $L_k := \{ i \in \{1,\ldots,m\} \;|\; k \in P_i\}$ as the set of all images containing the part $k$. For each part $k$ we can now define a constraint for the values that can take the variables $z_{k_j}$.

% \begin{equation}
%     \sum_{k = 1}^{|L_k|} z_{k_j}=|L_k|-1
% \end{equation}
\begin{equation}
    \sum_{j \in L_k} z_{k_j}=|L_k|-1
\end{equation}

The constraint expressed above states that only one of the $z_{k_j}$ variables can be 0, the rest have to be 1. We define the minimum resolution of a part as $r_k$. With the following two constraints, we can linearize (\ref{resolution-part-cp}).

\begin{equation}
    r_k \geq (x_jR_j + B(1-x_i)) -2Bz_{k_j} \text{ for all } j \in L_k
    \label{effective-res}
\end{equation}

\begin{equation}
    \min \sum_{k \in U} r_k
\end{equation}

The first term on the right-hand side of (\ref{effective-res}) affects how the part $k$ perceives the resolution of the images to which it belongs. If the image is in a cover, the resolution is equal to $R_j$. If the image is not in a cover, then the resolution is equal to the big constant $B$. As we minimize $r_k$, the lower this first term, the better. This term forced the images with lower resolution to be in a cover.
The second term on the right-hand side of (\ref{effective-res}) is used to force $r_k$ equal to the minimum value of the resolution of the images that contain the part $k$ and are in the cover.
When $z_{k_j} = 1$ the right-hand side is negative and when $0$ the value is positive, and it is the value that $r_k$ takes.  

To linearize the incidence angle objective, we need to minimize an auxiliary variable $max_f$ that represents the maximum incidence angle of the images in the cover.

\begin{equation}
    \min max_f \geq x_iF_i, \text{for all } i=1,\ldots,m
    \label{max-incidence-in-cover}
\end{equation}

To minimize the area of the clouds, we can model this as a partial set cover problem, where the universe $C = \{1, \ldots\, c\}$ is formed by all the clouds, and the sets are the images that can cover the clouds. For example, if we have the following set $P_{2_c} = \{c_1,c_2,c_5\}$ it means that image 2 can cover clouds 1, 2 and 5, i.e. parts 1, 2 and 5 are not cloudy in image 2. For each cloud $c_i$ we have a variable $y_i$ that is 1 if the cloud is covered or 0 otherwise and $A_c$ indicating the area of the cloud. To maximize the covering of the cloudy areas, we will minimize the following expression:

\begin{equation}
    \min-\sum_{c \in C} y_c A_c
\end{equation}

Subject to the following constraint, which forces $y_c$ to be $0$ if none of the images that cover the cloud $c$ is selected to cover the AOI.

\begin{equation}
    \sum_{i:c \in P_{i_c}} x_i \geq y_c , \text{for all $c \in C$}
\end{equation}

\section{MiniZinc Model}
\label{mzn-model}

We describe the full \textsc{MiniZinc} constraint model implementing the mathematical model given in Section~\ref{problem-spec}.

\begin{lstlisting}[language=minizinc,mathescape]
int: num_images;
int: universe;
int: max_cloud_area;

set of int: IMAGES = 1..num_images;
set of int: UNIVERSE = 1..universe;

array[IMAGES] of set of int: images;
array[IMAGES] of set of int: clouds;
array[IMAGES] of int: costs;
array[UNIVERSE] of int: areas;
array[IMAGES] of int: resolution;
array[IMAGES] of int: incidence_angle;

array[IMAGES] of var bool: taken;

% Which images have a universe `u` without cloud?
% That is, uclear[u] = {i1, i2, ..} means that the images numbered i1, i2, ... contains `u` without clouds.
array[UNIVERSE] of set of int: uclear = [{ i | i in IMAGES where not (u in clouds[i]) /\ u in images[i] } | u in UNIVERSE];

% Set covering constraint.
constraint forall(u in UNIVERSE)(
  exists(i in IMAGES)(taken[i] /\ u in images[i]));

% cloudy[u] is true iff no image containing a version of `u` without clouds is taken.
array[UNIVERSE] of var bool: cloudy;
array[UNIVERSE] of var int: num_clear_images;
constraint forall(u in UNIVERSE)(
  num_clear_images[u] = sum(i in uclear[u])(taken[i])
);
constraint forall(u in UNIVERSE)(cloudy[u] = (num_clear_images[u] == 0));

var int: cloudy_area = sum(u in UNIVERSE)(cloudy[u] * areas[u]);

var int: total_cost = sum(i in IMAGES)(costs[i] * taken[i]);
var int: max_resolution = sum(u in UNIVERSE)(min(i in IMAGES where u in images[i] /\ taken[i])(resolution[i]));
var int: max_incidence = max(i in IMAGES)(taken[i] * incidence_angle[i]);

array[1..4] of var int: objs;
constraint objs[1] = total_cost;
constraint objs[2] = cloudy_area;
constraint objs[3] = max_resolution;
constraint objs[4] = max_incidence;
\end{lstlisting}

\section{Experimental results detailed}
\label{experimental-result-detailed}
\begin{table}[H]
    \centering
    \caption{Hypervolume values for all the experiments.}
    \begin{tabular}{llllll}
        \hline
         Instance          & OR-tools default   & OR-tools greedy   & Gecode default    & Gecode greedy     & Gurobi            \\
        \hline
             lagos\_nigeria\_30   & \textbf{5.46E+33}  & 5.21E+33          & 4.89E+32          & 4.91E+33          & 4.87E+33          \\
         mexico\_city\_30     & 4.83E+32           & 4.82E+33          & 4.62E+33          & 4.59E+32          & \textbf{4.83E+33} \\
         paris\_30           & \textbf{1.95E+34}  & \textbf{1.95E+34} & 1.59E+34          & 1.58E+34          & 1.95E+33          \\
         rio\_de\_janeiro\_30  & \textbf{5.76E+33}  & 5.65E+33          & 5.65E+33          & 5.65E+33          & \textbf{5.76E+33} \\
         tokyo\_bay\_30       & \textbf{6.35E+33}  & 6.33E+33          & 6.05E+33          & 6.05E+33          & 6.30E+33          \\
         lagos\_nigeria\_50   & \textbf{3.26E+34}  & 3.15E+34          & 2.19E+34          & 2.18E+34          & 2.77E+34          \\
         mexico\_city\_50     & 2.88E+33           & \textbf{2.92E+34} & 2.63E+34          & 2.57E+34          & 2.71E+34          \\
         paris\_50           & \textbf{9.02E+34}  & 8.85E+34          & 7.12E+34          & 7.24E+34          & 7.85E+34          \\
         rio\_de\_janeiro\_50  & 3.94E+33           & \textbf{3.87E+34} & 3.05E+34          & 3.04E+34          & 3.83E+34          \\
         tokyo\_bay\_50       & \textbf{3.32E+34}  & 3.22E+34          & 1.65E+34          & 1.78E+34          & 1.98E+34          \\
         lagos\_nigeria\_100  & 1.86E+35           & \textbf{1.87E+35} & 1.68E+35          & 1.68E+35          & 8.55E+34          \\
         mexico\_city\_100    & 1.98E+35           & 1.97E+35          & \textbf{2.11E+35} & 2.03E+35          & 1.58E+35          \\
         paris\_100          & \textbf{4.73E+35}  & 3.13E+34          & 4.32E+35          & 2.91E+35          & 4.32E+35          \\
         rio\_de\_janeiro\_100 & \textbf{4.76E+35}  & 4.14E+35          & 2.28E+35          & 1.83E+35          & 3.39E+35          \\
         tokyo\_bay\_100      & 1.23E+35           & 1.23E+35          & 1.89E+35          & 1.89E+35          & \textbf{2.77E+35} \\
         lagos\_nigeria\_145  & 2.98E+36           & \textbf{3.78E+36} & 2.32E+36          & 2.32E+36          & 9.30E+35          \\
         mexico\_city\_150    & 1.52E+36           & 1.11E+36          & 2.27E+36          & \textbf{2.28E+36} & 3.42E+35          \\
         paris\_150          & 2.60E+36           & 2.83E+36          & 1.11E+36          & 1.11E+36          & \textbf{2.87E+36} \\
         rio\_de\_janeiro\_150 & \textbf{6.59E+35}  & 4.81E+34          & 4.00E+35          & 4.00E+35          & 4.61E+34          \\
         tokyo\_bay\_150      & \textbf{1.16E+36}  & 8.68E+35          & 8.36E+35          & 8.36E+35          & 1.12E+36          \\
         mexico\_city\_200    & 2.66E+36           & 2.05E+35          & \textbf{4.28E+36} & \textbf{4.28E+36} & 1.06E+36          \\
         paris\_200          & 0.00E+00           & 0.00E+00          & 0.00E+00          & 0.00E+00          & \textbf{1.60E+36} \\
         rio\_de\_janeiro\_200 & \textbf{0.00E+00}  & \textbf{0.00E+00} & \textbf{0.00E+00} & \textbf{0.00E+00} & \textbf{0.00E+00} \\
         tokyo\_bay\_200      & 0.00E+00           & 0.00E+00          & 0.00E+00          & 0.00E+00          & \textbf{2.46E+35} \\
        \hline
\end{tabular}
    \label{experimental-result-table}
\end{table}
\end{document}